\documentclass[nohyperref]{article}

\usepackage{microtype}
\usepackage{graphicx}
\usepackage{subfigure}
\usepackage{bm}
\usepackage{booktabs} 

\usepackage{hyperref}
\usepackage{lipsum}

\usepackage[accepted]{pre-training}

\usepackage{amsmath}
\usepackage{amssymb}
\usepackage{mathtools}
\usepackage{amsthm}

\usepackage[capitalize,noabbrev]{cleveref}
\usepackage{xspace}

\theoremstyle{plain}

\theoremstyle{definition}

\theoremstyle{remark}

\newcommand{\model}{\textsc{PiCo}\xspace}

\usepackage[textsize=tiny]{todonotes}

\icmltitlerunning{Pixel-level Correspondence for Self-Supervised Learning from Video}

\begin{document}

\twocolumn[
\icmltitle{Pixel-level Correspondence for Self-Supervised Learning from Video}



\icmlsetsymbol{intern}{$\dagger$}
\vspace{-0.5em}
\begin{icmlauthorlist}
\icmlauthor{Yash Sharma}{intern,tue}
\icmlauthor{Yi Zhu}{ama}
\icmlauthor{Chris Russell}{ama}
\icmlauthor{Thomas Brox}{ama,frei}
\end{icmlauthorlist}

\icmlaffiliation{tue}{University of T\"ubingen}
\icmlaffiliation{ama}{Amazon}
\icmlaffiliation{frei}{University of Freiburg}

\icmlcorrespondingauthor{Yash Sharma}{yash.sharma@bethgelab.org}

\vspace{1.5em}
]



\printAffiliationsAndNotice{\textsuperscript{$\dagger$}Work done during an internship at Amazon.} 
\begin{abstract}
While self-supervised learning has enabled effective representation learning in the absence of labels, for vision, video remains a relatively untapped source of supervision. To address this, we propose Pixel-level Correspondence (\model), a method for dense contrastive learning from video. By tracking points with optical flow, we obtain a correspondence map which can be used to match local features at different points in time. We validate \model on standard benchmarks, outperforming self-supervised baselines on multiple dense prediction tasks, without compromising performance on image classification. 
\end{abstract}

\section{Introduction}
Deep learning methods have yielded dramatic improvements in a plethora of domains by extracting useful representations from raw data ~\citep{bengio2013representation,lecun2015deep}, albeit assuming the availability of ample supervision. Recent advancements in self-supervised learning ~\citep{mikolov2013efficient,devlin2018bert,chen2020simple,he2021masked} have enabled effective representation learning without curated, labeled datasets~\citep{goyal2021self}.

Self-supervised learning obtains supervisory signals from the data itself through the careful construction of prediction tasks which do not rely on manual annotation, yet encourage the model to extract useful features. Specifically, the task of predicting whether a pair, or a set, of examples are views of the ``same'' image, or ``different'' images, underlies the recent success of contrastive methods for learning representations of visual data~\citep{wu2018unsupervised,van2018representation,henaff2020data,hjelm2018learning,bachman2019learning,he2020momentum,chen2020simple}. 

\begin{figure}[!t]
\centering
\includegraphics[width=0.75\linewidth]{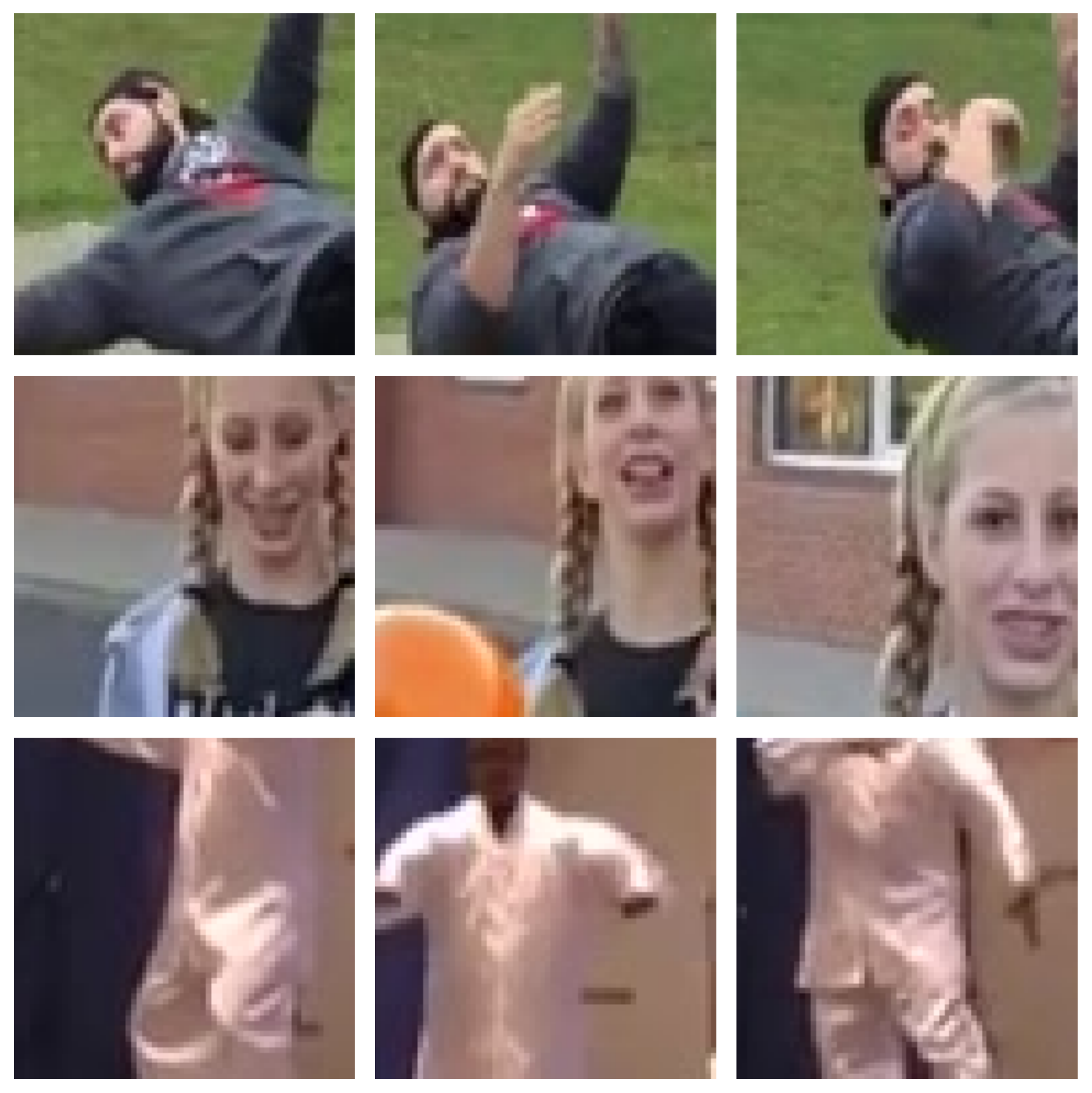}
\vspace{-1em}
\caption{Each row shows patches along point trajectories computed on Kinetics-400, resized for viewing. Clearly, hand-crafted transformations (crops, color distortion) used in practice cannot capture the variation depicted here.}
\label{fig:teaser}
\vspace{-2em}
\end{figure}

In contrastive learning, view selection crucially influences the quality of the resulting representations~\citep{tian2020makes,zimmermann2021contrastive,von2021self}. Existing approaches~\citep{he2020momentum,chen2020simple,chen2020improved} have constructed views via hand-crafted data augmentations, e.g. cropping sub-regions of the images. Cropping yields views that depict object parts, and thereby induces a learning signal for invariance to occluded objects~\citep{purushwalkam2020demystifying}. With that said, augmentations are inherently limited; given a single image, simulating variation in object size, shape, or viewpoint can be difficult. Notably, such variation is ubiquitous in video (see~\Cref{fig:teaser}). The promise of temporal variation for representation learning has encouraged ample investigation in the context of self-supervision~\citep{misra2016shuffle,wei2018learning,wang2019learning,vondrick2018tracking,isola2015learning,wiskott2002slow,klindt2020towards,agrawal2015learning,weis2021benchmarking,lachapelle2021disentanglement}.

How can we leverage video for learning self-supervised representations of images? While existing work has proposed a multitude of strategies~\citep{wang2015unsupervised,wang2017transitive,tschannen2020self,purushwalkam2020demystifying,gordon2020watching,romijnders2021representation,xiong2021self,wu2021contrastive,chen2021previts}, nearly all exploit instance discrimination methods~\citep{dosovitskiy2014discriminative,kolesnikov2019revisiting,he2020momentum,chen2020improved} designed for global representation learning, or learning encodings at the image-level. However, recent work~\citep{o2020unsupervised,wang2021dense,xie2021propagate,xiao2021region,bai2022point} has demonstrated that dense representation learning, or learning encodings at the region/pixel-level, can improve performance for dense prediction tasks (e.g. segmentation, depth prediction), at the cost of reduced performance for global prediction tasks (e.g. image classification)~\citep{xie2021propagate,xiao2021region}. Note that this observation echoes the related findings from empirical studies that ImageNet accuracy is not predictive for downstream tasks outside of image/scene classification~\citep{kotar2021contrasting,atanov2022simple}.

We thus propose \model, a method for dense representation learning from video. Existing work proposed for static images has relied upon aforementioned geometric transformations, e.g. crops, to introduce variation. We demonstrate that temporal variation can also be utilized by tracking points using off-the-shelf optical flow estimators. We find that across a number of downstream tasks, \model outperforms existing work restricted to static frames, as well as existing work applied to video assuming static pixel correspondence.  
\section{Background}
As our contribution enables dense representation learning to exploit the natural transformations inherent to video, we will focus on extending a method which learns representations through pixel-level contrastive learning, VADeR~\citep{o2020unsupervised}. Thus, we will give a short description of the learning method before proceeding with our contribution, see~\citep{o2020unsupervised} for further details.

Let us represent a pixel $u$ in image $\mathbf{x}\in\mathcal{I}\subset\mathbb{R}^{3\times h\times w}$ by the tuple $(\mathbf{x}, u)$. Let $f$ be an encoder-decoder convolutional network that produces a $d$-dimensional embedding for every pixel in the image, i.e. $f: (\mathbf{x}, u)\mapsto z \in \mathbb{R}^{d}$. VADeR's objective is to learn an embedding function that encodes $(\mathbf{x}, u)$ into a representation that is invariant w.r.t. any view $v_1,v_2\in\mathcal{V}_u$ containing the pixel $u$. This is achieved through contrastive learning~\citep{gutmann2010noise,gutmann2012noise,van2018representation}, where the objective optimized in practice is to distinguish between views of the same pixel and views of different pixels. 
\begin{equation}
\label{eq:InfoNCE_objective}
\textstyle
\mathcal{L}_{\text{InfoNCE}}=\\
-\mathbb{E}_{(v_1,v_2)\sim\mathcal{V}_u}\Big[\log \frac{\exp\{\text{sim}(f(v_1,u),f(v_2,u))\}}{\sum_{j=1}^K\exp\{\text{sim}(f(v_1,u),f(v'_j,u'_j))\}
}\Big]
\end{equation}
where $K-1$ is the number of negative pixels, and the positive pair in the numerator is included in the denominator summation, i.e. $(v'_K,u'_K)=(v_2,u)$ . For implementation, the design details for MoCo were followed~\citep{he2020momentum}. 

For $f$, the semantic segmentation branch of~\citep{kirillov2019panoptic} was adopted. A feature pyramid network (FPN)~\citep{lin2017feature} adds a top-down path to a ResNet-50~\citep{he2016deep}, generating a pyramid of features (from $1/32$ to $1/4$ resolution). By adding a number of upsampling blocks at each resolution of the pyramid, the pyramid representations are merged into a single dense output representation with dimension $128$ and scale $1/4$. The ResNet-50 is initialized with MoCo~\citep{he2020momentum}, and pretraining is performed on the ImageNet-1K (IN-1K)~\citep{deng2009imagenet} train split. 
\section{Method}
Here, we detail our procedure for constructing pixel correspondence maps from video for dense contrastive learning.

\subsection{Data} 
For pretraining, we experiment with Kinetics400 (K400)~\citep{kay2017kinetics} and YouTube-8M (YT8M)~\citep{abu2016youtube}. The K400 training set consists of approximately 240,000 videos trimmed to $10$ seconds from $400$ human action categories. We sample frame sequences at 30 Hz~\citep{kuang2021video}. For tractability, we construct a subset of YT8M (YT8M-S) which matches the dataset statistics of K400. Specifically, for 240,000 random videos, we sample $10$-second snippets at 30 Hz from shots detected using an off-the-shelf network~\citep{souvcek2020transnet}. Further details are provided in the appendix. 

\subsection{Trajectories} 
We first compute and store optical flow on K400 and YT8M-S. While in preliminary experiments we found alternatives~\citep{ilg2017flownet,sun2018pwc} to perform comparably, for the presented set of experiments, we use RAFT~\citep{teed2020raft} trained on a mixed dataset~\citep{2021mmflow} consisting of FlyingChairs~\citep{dosovitskiy2015flownet}, FlyingThings3D~\citep{mayer2016large}, Sintel~\citep{butler2012naturalistic}, KITTI-2015~\citep{menze2015object,geiger2013vision}, and HD1K~\citep{kondermann2016hci}. The horizontal and vertical components of the flow were 
linearly rescaled to a $[0,255]$ range and compressed using JPEG (after decompression, the flow is rescaled back to its original range)~\citep{simonyan2014two}.

With the precomputed flow, we track points in the video. For each video, we sample an initial set of $1000$ points at random locations on random frames. As in~\citep{sundaram2010dense}, each point is tracked to the next frame using the flow field $\mathbf{w}=(u,v)^{T}$:
\begin{equation}
\label{eq:tracking}
\textstyle
(x_{t+1},y_{t+1})^{T}=(x_{t},y_{t})^{T}+(u_{t}(x_{t},y_{t}),v_{t}(x_{t},y_{t}))^{T}
\end{equation}
Between pixels, the flow is inferred using bilinear interpolation. Tracking is stopped as soon as a point is occluded, which is detected by checking the consistency of the forward and backward flow. In a non-occlusion case, the backward flow vector should point in the inverse direction
of the forward flow vector: $u_{t}(x_{t},y_{t})=-\hat{u}_{t}(x_{t}+u_{t},y_{t}+v_{t})$ and $v_{t}(x_{t},y_{t})=-\hat{v}_{t}(x_{t}+u_{t},y_{t}+v_{t})$, where $\mathbf{\hat{w}}_{t}=(\hat{u}_{t},\hat{v}_{t})$ denotes the flow from frame $t+1$ to frame $t$. We thus use the following threshold:
\begin{equation}
\label{eq:consistency}
\textstyle
|\mathbf{w}+\mathbf{\hat{w}}|^{2}<\gamma(|\mathbf{w}|^{2}+|\mathbf{\hat{w}}|^{2})+\delta
\end{equation}
\subsection{Learning} 
Existing proposals for visual representation learning with contrastive methods from video typically sample random frames from a given shot for constructing views~\citep{tschannen2020self,chen2021previts,gordon2020watching}. Given the endpoints for a set of trajectories in each video, we propose a frame selection strategy for maximizing temporal separation and trajectory density, \textbf{anchor sampling}. After sampling an anchor frame, for each trajectory active on said frame, we find the endpoint furthest from said frame. If we are to select $N$ frames for learning, we select the top-$N$ according to endpoint count. With this strategy, as we vary the threshold hyperparameters, the temporal separation between the selected frames varies accordingly. 

\subsection{Implementation Details}
For both implementing the objective and initializing the encoder, we use MoCo-v2~\citep{chen2020improved,chen2020simple} instead of MoCo~\citep{he2020momentum}. Notably, we found the use of a nonlinear projection head to be critical for performance. As in existing work~\citep{long2015fully,wang2021dense,bai2022point}, for dense contrastive learning, we replace the linear layers in the MoCo-v2 global projection head with identical $1\times1$ convolution layers. Note that we use $2048$-dimensional hidden layers for the projector in concert with the 128-dimensional dense output representation; we found that reducing the number of parameters in the dense projection head decreased downstream performance. We also decided to freeze the initialized encoder, thereby maintaining the downstream image/scene classification performance of the image-level encoding. 

Finally, note that for the experiments prior to ablation, $\gamma=0$, $\delta=4.0$, and, each iteration, no more than $65536$ point pairs (from frame pairs selected from $256$ videos) are used. 
\section{Experiments}

\begin{figure}[!t]
\centering
\includegraphics[width=\linewidth]{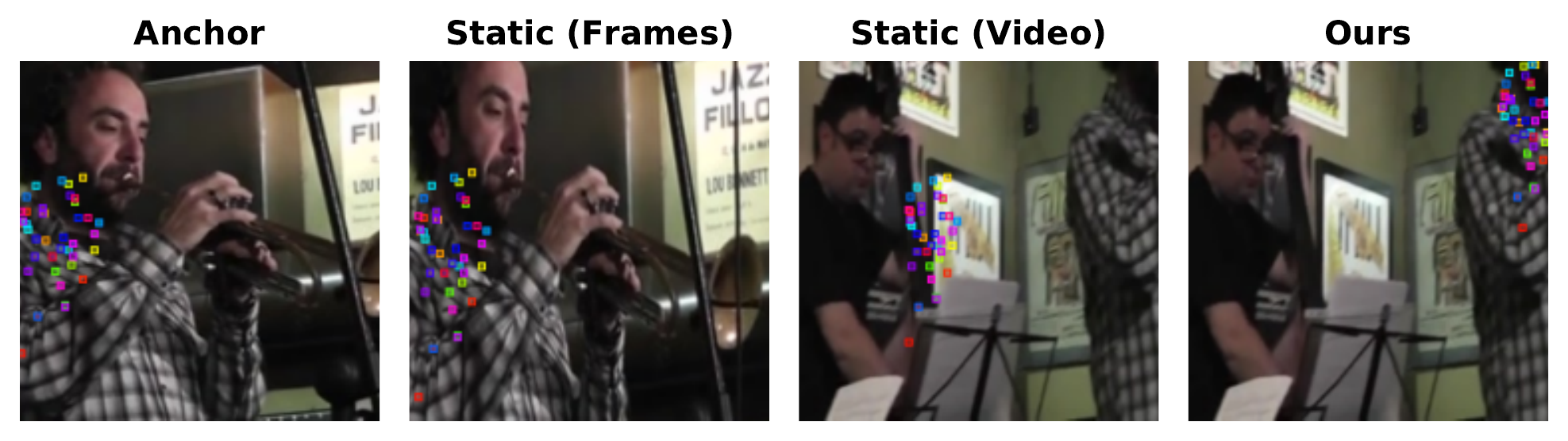}
\vspace{-1.5em}
\caption{For \textbf{Static (Frames)}, variation between local features stems solely from geometric transformations, e.g. random crops. For video, without point tracking, the static pixel correspondence map used by \textbf{Static (Video)} becomes imprecise with increased temporal separation. In contrast, we leverage off-the-shelf optical flow estimators to match local features over time.}
\vspace{-0.5em}
\label{fig:baselines}
\end{figure}

We compare \model to a set of baselines across datasets \& tasks. We provide a visual representation of the comparison in~\Cref{fig:baselines}. In \textbf{Static (Frames)}, a single frame is sampled from each video for view construction, thus, as in~\citep{o2020unsupervised}, the variation in corresponding pixels is solely due to the geometric transformations in the MoCo-v2 data augmentation pipeline, i.e. random crops and horizontal flips. In \textbf{Static (Video)}, as in \model, we sample multiple frames from a given video, but unlike \model, the pixel correspondence map is static, i.e. the optical flow field is assumed to consist of zero vectors. By comparing to ``Static (Video)'', we can isolate the value point tracking is yielding downstream. Additional details regarding evaluation are provided in the appendix. 
\subsection{COCO Semantic Segmentation}
It is common practice in self-supervised learning to assess the quality of frozen features with a \textbf{linear probe}~\citep{goyal2019scaling,kolesnikov2019revisiting}. Following~\citep{o2020unsupervised}, the output of each model is processed by a a $1\times1$ convolutional layer, $4\times$ upsample, and softmax, where for MoCo-v2, the effective stride is reduced from $1/32$ to $1/4$ by replacing strided convolutions with dilated ones~\citep{chen2017deeplab,yu2017dilated}. The linear predictor weights are trained using cross entropy. 

\begin{table}
\vspace{-0.5em}
\centering
\caption{\textbf{Linear probe.} COCO semantic segmentation.}
\label{table:coco}
\small
\begin{tabular}{ll|rr}
\toprule
Method & Dataset & mIoU & fIoU \\
\toprule
MoCo-v2 (IN-1K) & & 11.2 & 39.9 \\
\midrule
Static (Frame) & K400 & 18.4 & 49.5 \\
Static (Video) & K400 & 18.7 & 49.5 \\
\model & K400 & \textbf{20.9} & \textbf{51.6} \\
\midrule
Static (Frame) & YT8M-S & 17.0 & 48.3 \\
Static (Video) & YT8M-S & 18.3 & 49.6 \\
\model & YT8M-S & \textbf{19.9} & \textbf{50.5} \\
\bottomrule
\end{tabular}
\vspace{-0.5em}
\end{table}

In~\Cref{table:coco}, we observe a tangible improvement in leveraging point trajectories for dense contrastive learning. Interestingly, we find pretraining on K400 largely delivers improved performance relative to YT8M-S. In accordance with previous work~\citep{gordon2020watching}, we notice that a number of videos in YT8M are unnatural, e.g. ``video games'' or ``cartoons'', which clearly yields a domain gap with ``everyday scenes containing common objects in their natural context''~\citep{lin2014microsoft}.  

\subsection{Additional Tasks \& Benchmarks}

\begin{table}
\vspace{-0.5em}
\centering
\caption{\textbf{Mask R-CNN}. K400 pretraining, COCO object detection \& instance segmentation with FPN frozen.}
\label{table:coco3}
\small
\resizebox{\linewidth}{!}{%
\begin{tabular}{ll|rrr|rrr}
\toprule
Method & Dataset & AP$^{\text{bb}}$ & AP$^{\text{bb}}_{50}$ & AP$^{\text{bb}}_{75}$ & AP$^{\text{mk}}$ & AP$^{\text{mk}}_{50}$ & AP$^{\text{mk}}_{75}$ \\
\midrule
Static (Frame) & K400 & 6.09 & 15.1 & 3.73 & 7.37 & 14.8 & 6.64\\
Static (Video) & K400 & 7.92 & 19.7 & 4.60 & 9.06 & 18.7 & 7.72\\
\model & K400 & \textbf{10.8} & \textbf{24.3} & \textbf{7.94} & \textbf{11.9} & \textbf{23.1} & \textbf{10.9}\\
\bottomrule
\end{tabular}%
}
\vspace{-0.5em}
\end{table}

\paragraph{Tasks:}
In~\Cref{table:coco3}, we evaluate representations on COCO object detection and instance segmentation. For this, we use Mask R-CNN~\citep{he2017mask} with a frozen FPN backbone~\citep{lin2017feature}. While \model significantly outperforms the baselines, without being able to adapt the backbone downstream, the absolute scores are low. 

\begin{table}
\vspace{-0.5em}
\centering
\caption{\textbf{Additional Benchmarks}: K400 pretraining, linear probing frozen model. }
\label{table:coco4}
\small
\begin{tabular}{c|cc|c}
\toprule
& \multicolumn{2}{c}{sem. seg. (mIoU)}  & depth (RMSE)\\
Method & VOC & CS & NYU-d v2\\
\midrule
Static (Frame) & 31.0 & 34.0 & 1.000 \\
Static (Video) & 34.7 & 28.7 & 0.958 \\
\model & \textbf{35.6} & \textbf{35.1} & \textbf{0.950}\\
\bottomrule
\end{tabular}
\vspace{-0.5em}
\end{table}

\paragraph{Benchmarks:}
In~\Cref{table:coco4}, we evaluate on two additional datasets for semantic segmentation (Pascal VOC 2012~\citep{everingham2010pascal} and Cityscapes~\citep{cordts2016cityscapes}), as well as a dataset for depth prediction (NYU-depth v2~\citep{silberman2012indoor}). For depth prediction, we no longer apply the softmax function, and instead minimize the $L_1$ loss between the linear output and the per-pixel ground-truth depth values~\citep{kotar2021contrasting,o2020unsupervised}. We find that across all tasks, \model outperforms the baselines.  
\subsection{Ablations}
\begin{figure}[!t]
\centering
\includegraphics[width=0.5\linewidth]{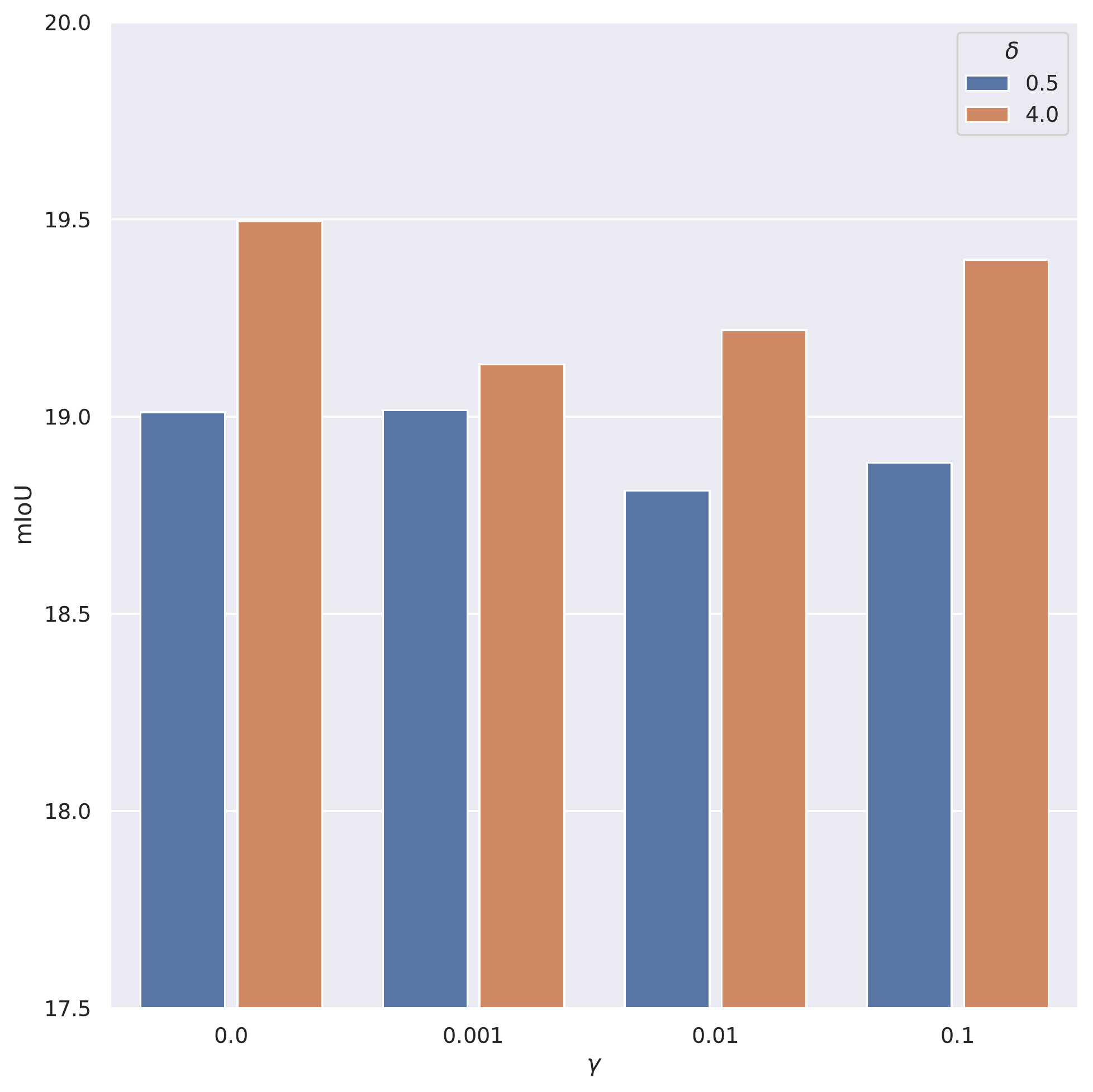}
\vspace{-1em}
\caption{\textbf{$\boldsymbol{\gamma}$:} COCO linear probing varying the tracking threshold parameters $\gamma$ and $\delta$. YT-8M-S pretraining, w/o anchor sampling.}
\vspace{-1em}
\label{fig:fig2}
\end{figure}
\begin{figure}[!t]
\centering
\includegraphics[width=0.5\linewidth]{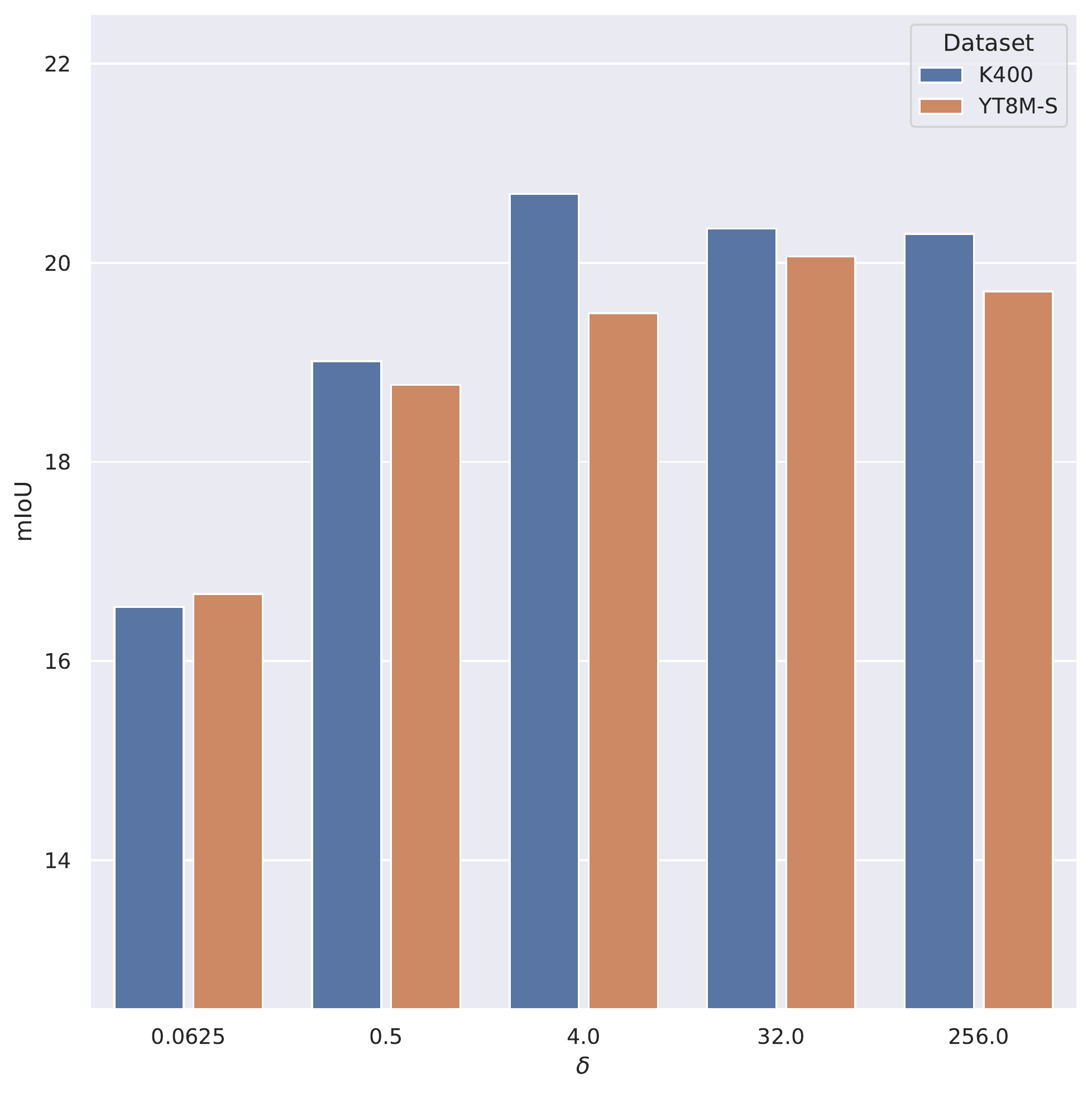}
\vspace{-1em}
\caption{$\boldsymbol{\delta}$: COCO linear probing with finer-grained variation in $\delta$. $\gamma=0$, w/o anchor sampling.}
\vspace{-1em}
\label{fig:fig3}
\end{figure}
\paragraph{Tracking Threshold:} 
In~\Cref{fig:fig2}, we evaluate the impact of varying $\gamma$ and $\delta$. While we do consistently observe improved performance with increased $\delta$ (up to a point, see~\Cref{fig:fig3}), the same cannot be said for $\gamma$. Given the computational cost in adjusting the trajectories w.r.t. $\gamma$, we were limited in our ablation, and encourage further exploration on the effect of this parameter. 

\begin{figure}[!t]
\centering
\includegraphics[width=0.5\linewidth]{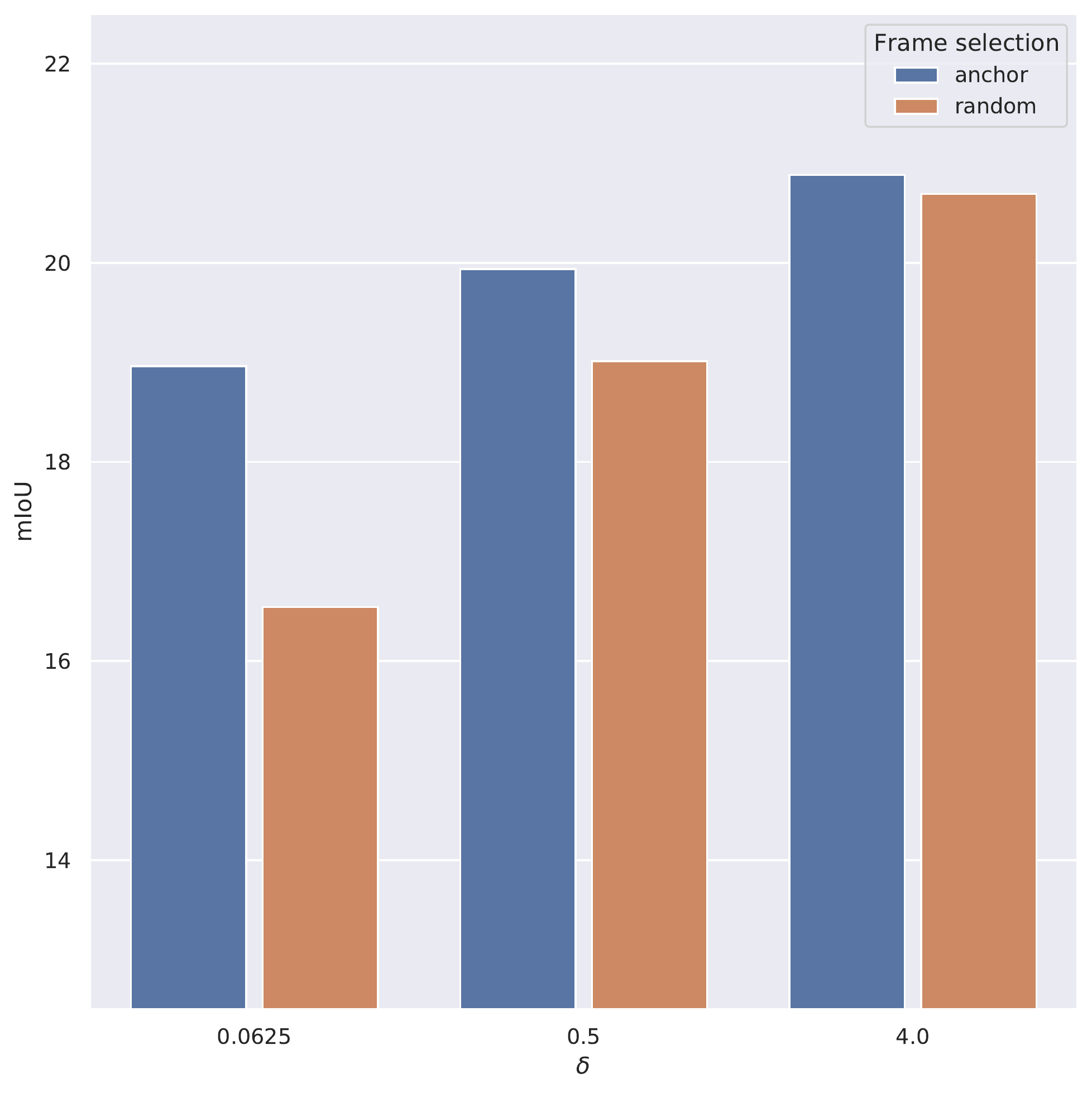}
\vspace{-1em}
\caption{\textbf{Anchor Sampling:} COCO linear probing w/ and w/o the anchor sampling strategy. K400 pretraining.}
\vspace{-1em}
\label{fig:fig4}
\end{figure}

\paragraph{Anchor Sampling:}
In~\Cref{fig:fig4}, we isolate the effect of anchor sampling on the downstream performance. We can see that as we increase $\delta$, thereby using longer trajectories for pretraining, the gap between anchor sampling and randomly sampling frames narrows. As $\delta$ increases, the likelihood that a point pair will exist between a random pair of frames also increases, while the very same likelihood is invariant to $\delta$ when using anchor sampling.  

\section{Discussion}
\paragraph{Limitations} While we observe improved performance over our baseline methods, overall performance remains worse than supervised approaches, and there is substantial room for improvement. Specifically, our decoder-only training on video, when compared to the reported scores of encoder-decoder training on IN-1K in a similar experimental setting~\citep{o2020unsupervised}, underperforms. In future work, we suggest (i) addressing the domain gap between the video datasets used for pretraining and the image datasets used for benchmarking~\citep{tang2012shifting,kalogeiton2016analysing,kae2020image} and (ii) considering alternatives to our strategy of freezing the encoder for maintaining classification performance whilst improving dense prediction. 

\paragraph{Conclusion}
We present \model, an approach to dense contrastive learning on video. We show the benefit in constructing pixel correspondence maps over time on a number of tasks and datasets. Our work serves as a first step towards leveraging the temporal variation inherent to video for dense prediction tasks, and in that vein, we encourage further exploration along the aforementioned direction.

\section*{Acknowledgements}
We thank Peter Gehler, Andrii Zadaianchuk, and Max Horn for compute infrastructure support. This work was supported by the German Federal Ministry of Education and Research (BMBF): Tübingen AI Center, FKZ: 01IS18039A. Y.S. thanks the International Max Planck Research School for Intelligent Systems (IMPRS-IS) for support.

\bibliography{example_paper}
\bibliographystyle{icml2022}

\newpage
\appendix
\onecolumn
\section{Additional Details}
\subsection{Training}
\paragraph{YT8M-S} Given the size of YT-8M, the authors decided to release frame-level features of the videos instead of the videos themselves~\citep{abu2016youtube}. For our purposes, we extracted YT-8M URLs\footnote{used following   \href{https://github.com/danielgordon10/youtube8m-data}{repository}.}, and downloaded a sampled subset at the scale of K-400. We note that a number of the YT-8M URLs are no longer accessible. We used TransNetV2~\citep{souvcek2020transnet} off-the-shelf as a high-performing deep learning approach for shot boundary detection.
\paragraph{Tracking} The most notable difference with~\citep{sundaram2010dense} corresponds to starting point sampling. In~\citep{sundaram2010dense}, a grid is instantiated on the first frame, and points are re-instantiated as trajectories are stopped. In contrast, we sampled starting points uniformly in space and time, to ensure the same trajectory computation is applicable to variable $\gamma$ and $\delta$. For storing the trajectories, in particular the consecutive norm differences between forward and backward flow vectors, we used half-precision. Finally, note that the RHS of~\Cref{eq:consistency} is dependent on the flow vectors through the $\gamma$ term, thus tuning $\gamma$ requires extra computation relative to solely tuning $\delta$. 
\paragraph{Training} In order to compute the loss, we must map point pairs to feature indices. For this, we simply scale the point indices by $1/4$, given the dense output representation is at $1/4$ resolution.  
\subsection{Evaluation}
For each configuration, we use the default FPN config provided in \texttt{Detectron2}\footnote{see \href{https://github.com/facebookresearch/detectron2/blob/main/configs/Base-RCNN-FPN.yaml}{here}} as a basis.
\subsubsection{COCO Semantic Segmentation}
\paragraph{Dataset:} Following~\citep{kirillov2019panoptic}, semantic annotations are converted from panoptic annotations for the 2017 challenge images, where all ``things'' are assigned the same semantic label, while each ``stuff'' category is assigned a unique semantic label. 
\paragraph{MoCo-v2:} As in~\citep{o2020unsupervised}, the dilated resnet architecture is used~\citep{chen2017deeplab,yu2017dilated}. For each stage where the stride is decreased from $2$ to $1$, the dilation factor is multiplicatively scaled by $2$. With that, the output resolution of the RN-50 is $1/4$, and can thereby be evaluated using the same linear prediction protocol as used for the encoder-decoder architectures. 
\paragraph{Configuration:} For data augmentation, we perform random absolute crops of size $672\times672$ after resizing using the default parameters, followed by a random flip. 
\subsubsection{COCO Instance Segmentation \& Object Detection}
\paragraph{Configuration:} Only discrepancy with the default configuration is freezing the FPN. Thus, in contrast to the semantic segmentation \& depth prediction evaluation, where solely a linear predictor is learned, the learned modules here are the proposal generator \& ROI heads. 
\subsubsection{VOC \& Cityscapes Semantic Segmentation}
\paragraph{VOC Configuration:} The minimum size after resizing was decreased to $480$, and an absolute crop size of $512\times512$ was specified. Number of gradient steps was decreased to $40000$, with milestone steps decreased to $25000$ and $35000$. Note that training was performed on the ``train\_aug'' dataset. 
\paragraph{Cityscapes Configuration:} The minimum size after resizing was decreased to $512$, and the maximum size was increased to $2048$. Crops of size $512\times1024$ were performed. Batch size was increased from $16$ to $32$, base learning rate was decreased from $0.02$ to $0.01$, and the number of gradient steps was decreased to $65000$, with milestone steps decreased to $40000$ and $55000$. 
\subsubsection{NYU-Depth V2 Depth Prediction}
\paragraph{Dataset:} Downloaded from the ViRB framework release\footnote{see \href{https://github.com/allenai/ViRB\#dataset-download}{here}}~\citep{kotar2021contrasting}. 
\paragraph{Configuration:} Given the variable resolution, both input examples and labels were resized to $224\times224$ prior to training and testing. For augmentation, only random flips were employed.

\end{document}